\documentclass[11pt,letterpaper]{article}

\makeatletter
\newcommand{\@BIBLABEL}{\@emptybiblabel}
\newcommand{\@emptybiblabel}[1]{}
\makeatother
\usepackage{hyperref}

\usepackage{eacl2017}
\usepackage{latexsym}
\usepackage{times}
\usepackage{xspace}
\usepackage{graphicx}

\usepackage[normalem]{ulem}

\eaclfinalcopy

\def\eaclpaperid{185}

\newcommand{\txt}[1]{\textsf{#1}}
\newcommand{\wiki}[1]{\texttt{#1}}

\newcommand{\kb}{\textsc{kb}\xspace}
\newcommand{\kbs}{\textsc{kb}s\xspace}
\newcommand{\lstm}{\textsc{lstm}\xspace}

\newcommand{\rnn}{\textsc{rnn}\xspace}
\newcommand{\rnns}{\textsc{rnn}s\xspace}
\newcommand{\nlm}{\textsc{nlm}\xspace}

\newcommand{\bleu}{\textsc{bleu}\xspace}

\newcommand{\train}{\textsc{train}\xspace}
\newcommand{\dev}{\textsc{dev}\xspace}
\newcommand{\test}{\textsc{test}\xspace}

\newcommand{\wikitext}{\textsc{wiki}\xspace}
\newcommand{\base}{\textsc{base}\xspace}
\newcommand{\stos}{\textsc{s2s}\xspace}
\newcommand{\stosae}{\textsc{s2s+ae}\xspace}

\title{Learning to generate one-sentence biographies from Wikidata}
\author{
    Andrew Chisholm\\
    University of Sydney\\
    Sydney, Australia\\
    {\tt \small{andy.chisholm.89@gmail.com}}
	\And
	Will Radford\\
    Hugo Australia\\
    Sydney, Australia\\
    {\tt \small{wradford@hugo.ai}}
	\And
	Ben Hachey\\
    Hugo Australia\\
    Sydney, Australia\\
  {\tt \small{bhachey@hugo.ai}}}
\date{}

\begin{document}

\maketitle

\begin{abstract}
We investigate the generation of one-sentence Wikipedia biographies from facts derived from Wikidata slot-value pairs.
We train a recurrent neural network sequence-to-sequence model with attention to select facts and generate textual summaries.
Our model incorporates a novel secondary objective that helps ensure it generates sentences that contain the input facts.
The model achieves a \bleu score of 41, improving significantly upon the vanilla sequence-to-sequence model and scoring roughly twice that of a simple template baseline.
Human preference evaluation suggests the model is nearly as good as the Wikipedia reference.
Manual analysis explores content selection, suggesting the model can trade the ability to infer knowledge against the risk of hallucinating incorrect information.
\end{abstract}

\section{Introduction}

Despite massive effort, Wikipedia and other collaborative knowledge bases (\kbs) have coverage and quality problems.
Popular topics are covered in great detail, but there is a long tail of specialist topics with little or no text.
Other text can be incorrect, whether by accident or vandalism.
We report on the task of generating textual summaries for people, mapping slot-value facts to one-sentence encyclopaedic biographies.
In addition to initialising stub articles with only structured data, the resulting model could be used to improve consistency and accuracy of existing articles.
Figure \ref{wikidata} shows a Wikidata
entry for \wiki{Mathias Tuomi}, with fact keys and values flattened into a sequence, and the first sentence from his Wikipedia
article.
Some values are in the text, others are missing (e.g. \txt{male}) or expressed differently (e.g. dates).

\begin{figure}[t!]
  \centering
  \texttt{
    \begin{tabular}{p{7.5cm}}
    \hline
    TITLE mathias tuomi SEX\_OR\_GENDER male DATE\_OF\_BIRTH 1985-09-03 \\
    OCCUPATION squash player CITIZENSHIP finland \\
    \hline
    \end{tabular}
  }
  \caption{Example Wikidata facts encoded as a flat input string. The first sentence of the Wikipedia article reads: \txt{Mathias Tuomi, (born September 30, 1985 in Espoo) is a professional squash player who represents Finland.}}
  \label{wikidata}
\end{figure}

We treat this \emph{knowlege-to-text} task like translation, using a recurrent neural network (\rnn) sequence-to-sequence model \cite{DBLP:conf/nips/SutskeverVL14} that learns to select and realise the most salient facts as text.
This includes an attention mechanism to focus generation on specific facts, a shared vocabulary over input and output, and a multi-task autoencoding objective for the complementary extraction task.
We create a reference dataset comprising more than 400,000 knowledge-text pairs, handling the 15 most frequent slots.
We also describe a simple template baseline for comparison on \bleu and crowd-sourced human preference judgements over a heldout \test set.

Our model obtains a \bleu score of 41.0, compared to 33.1 without the autoencoder and 21.1 for the template baseline.
In a crowdsourced preference evaluation, the model outperforms the baseline and is preferred 40\% of the time to the Wikipedia reference.
Manual analysis of content selection suggests that the model can infer knowledge but also makes mistakes, and that the autoencoding objective encourages the model to select more facts without increasing sentence length.
The task formulation and models are a foundation for text completion and consistency in \kbs.

\section{Background}

\rnn sequence-to-sequence models \cite{DBLP:conf/nips/SutskeverVL14} have driven various recent advances in natural language understanding.
While initial work focused on problems that were sequences of the same units, such as translating a sequence of words from one language to another, other work been able to use these models by \emph{coercing} different structures into sequences, e.g., flattening trees for parsing \cite{NIPS2015_5635}, predicting span types and lengths over byte input \cite{gillick-EtAl:2016:N16-1} or flattening logical forms for semantic parsing \cite{xiao-dymetman-gardent:2016:P16-1}.

\rnn{}s have also been used successfully in \emph{knowledge-to-text} tasks for human-facing systems, e.g., generating conversational responses \cite{DBLP:journals/corr/VinyalsL15}, abstractive summarisation \cite{rush-chopra-weston:2015:EMNLP}.
Recurrent \lstm models have been used with some success to generate text that completely expresses a set of facts: restaurant recommendation text from dialogue acts \cite{wen-EtAl:2015:EMNLP}, weather reports from sensor data and sports commentary from on-field events \cite{mei2015selective}.
Similarly, we learn an end-to-end model trained over key-value facts by flattening them into a sequence.

Choosing the salient and consistent set of facts to include in generated output is also difficult.
Recent work explores unsupervised autoencoding objectives in sequence-to-sequence models, improving both text classification as a pretraining step \cite{DBLP:conf/nips/DaiL15} and translation as a multi-task objective \cite{DBLP:journals/corr/LuongLSVK15}.
Our work explores an autoencoding objective which selects content as it generates by constraining the text output sequence to be predictive of the input.

Biographic summarisation has been extensively researched and is often approached as a sequence of subtasks \cite{schiffman-mani-concepcion:2001:ACL}.
A version of the task was featured in the Document Understanding Conference in 2004 \cite{Blair:2004} and other work learns policies for content selection without generating text \cite{Duboue-McKeown:2003:EMNLP,Zhang:2012:SHS:2348283.2348306,Cheng:2015:SED:2736277.2741094}.
While pipeline components can be individually useful, integrating selection and generation allows the model to exploit the interaction between them.

\kbs have been used to investigate the interaction between structured facts and unstructured text.
Generating textual templates that are filled by structured data is a common approach and has been used for conversational text \cite{han-EtAl:2015:W15-46} and biographical text generation \cite{duma-klein:2013:IWCS2013}.
Wikipedia has also been a popular resource for studying biography, including sentence harvesting and ordering \cite{biadsy-hirschberg-filatova:2008:ACLMain}, unsupervised discovery of distinct sequences of life events \cite{DBLP:journals/tacl/BammanS14} and fact extraction from text \cite{garera-yarowsky:2009:EACL}.
There has also been substantial work in generating from other structured \kbs using template induction \cite{kondadadi-howald-schilder:2013:ACL2013}, semantic web techniques \cite{power-third:2010:POSTERS}, tree adjoining grammars  \cite{gyawali-gardent:2014:P14-1}, probabilistic context free grammars \cite{konstas-lapata:2012:ACL2012} and probabilistic models that jointly select and realise content \cite{angeli-liang-klein:2010:EMNLP}.

\newcite{LeBret16} present the closest work to ours with a similar task using Wikipedia infoboxes in place of Wikidata.
They condition an attentional neural language model (\nlm) on local and global properties of infobox tables, including \emph{copy actions} that allow wholesale insertion of values into generated text.
They use 723k sentences from Wikipedia articles with 403k lower-cased words mapping to 1,740 distinct facts.
They compare to a 5-gram language-model with copy actions, and find that the \nlm has higher \bleu and lower perplexity than their baseline.
In contrast, we utilise a deep recurrent model for input encoding, minimal slot value templating and greedy output decoding.
We also explore a novel autoencoding objective that measures whether input facts can be re-created from the generated sentence.

Evaluating generated text is challenging and no one metric seems appropriate to measure overall performance.
\newcite{LeBret16} report \bleu scores \cite{Papineni:2002:BMA:1073083.1073135} which calculate the n-gram overlap between text produced by the system with respect to a human-written reference.
Summarisation evaluations have concentrated on the content that is included in the summary, with semantic content typically extracted manually for comparison \cite{lin-naacl03-automatic,nenkova-passonneau:2004:HLTNAACL}.
We draw from summarisation and generation to formulate a comprehensive evaluation based on automated metrics and human validation.
Our final system comparison follows \newcite{kondadadi-howald-schilder:2013:ACL2013} in running a crowd task to collect pairwise preferences for evaluating and comparing both systems and references.

\section{Task and Data}
\label{section:task}

We formulate the one-sentence biography generation task as shown in Figure \ref{wikidata}.
Input is a flat string representation of the structured data from the \kb, comprising slot-value pairs (the subject being the topic of the \kb record, e.g., \textsf{Mathias Tuomi}), ordered by slot frequency from most to least common.
Output is a biography string describing the salient information in one sentence.

We validate the task and evaluation using a closely-aligned set of resources: Wikipedia and Wikidata.
In addition to the \kb maintenance issues discussed in the introduction, Wikipedia first sentences are of particular interest because they are clear and concise biographical summaries.
These could be applied to entities outside Wikipedia for which one can obtain comparable parallel structured/textual data, e.g., movie summaries from IMDb, resume overviews from LinkedIn, product descriptions from Amazon.

\begin{table}[t!]
  \centering
  \begin{tabular}{lrr}
  \hline
  Fact & Count & \% \\
  \hline \hline
  TITLE (name)	&	1,011,682 & 98\\
  SEX\_OR\_GENDER	&	1,007,575 & 0\\
  DATE\_OF\_BIRTH	&	817,942 & 88 \\
  OCCUPATION	&	720,080 & 67\\
  CITIZENSHIP	&	663,707 & 52\\
  DATE\_OF\_DEATH	&	346,168 & 86 \\
  PLACE\_OF\_BIRTH	&	298,374 & 25\\
  EDUCATED\_AT &   141,334 &   32\\
  SPORTS\_TEAM	&	108,222 & 29\\
  PLACE\_OF\_DEATH &   107,188 &   17\\
  POSITION\_HELD   &   87,656  &   75\\
  PARICIPANT\_OF   &   77,795  &   23\\
  POLITICAL\_PARTY &   74,371  &   49\\
  AWARD\_RECEIVED  &   67,930  &   44\\
  SPORT    &   36,950  &   72\\
  \hline
  \end{tabular}
  \caption{The top fifteen slots across entities used for input, and the \% of time the value is a substring in the entity's first sentence.}
  \label{relations}
\end{table}

We use snapshots of Wikidata (2015/07/13) and Wikipedia (2015/10/02)
and batch process them to extract instances for learning.
We select all entities that are \texttt{INSTANCE\_OF human} in Wikidata.
We then use \texttt{sitelinks} to identify each entity's Wikipedia article text and \textsc{nltk} \cite{BirdKleinLoper09} to tokenize and extract the lower-cased first sentence.
This results in 1,268,515 raw knowledge-text pairs.
The summary sentences can be long and the most frequent length is 21 tokens.
We filter to only include those between the 10th and 90th percentiles: 10 and 37 tokens.
We split this collection into \train, \dev and \test collections with 80\%, 10\% and 10\% of instances allocated respectively.
Given the large variety of slots which may exist for an entity, we restrict the set of slots used to the top-15 by occurrence frequency.
This criteria covers 72.8\% of all facts. % (6163532.0 / 8462376.0)
Table \ref{relations} shows the distribution of fact slots in the structured data and the percentage of time tokens from a fact value occur in the corresponding Wikipedia summary.

Additionally, some Wikidata entities remain underpopulated and do not contain sufficient facts to reconstruct a text summary.
We control for this information mismatch by limiting our dataset to include only instances with at least 6 facts present.
The final dataset includes 401,742 \train, 50,017 \dev and 50,030 \test instances.
Of these instances, 95\% contain 6 to 8 slot values while 0.1\% contain the maximum of 10 slots.
51\% of unique slot-value pairs expressed in \test and \dev are not observed in \train so generalisation of slot usage is required for the task.
The \kb facts give us an opportunity to measure the correctness of the generated text in a more precise way than text-to-text tasks.
We use this for analysis in Section \ref{sec:analysis}, driving insight into system characteristics and implications for use.

\subsection{Task complexity}

Wikipedia first sentences exhibit a relatively narrow domain of language in comparison to other generation tasks such as translation.
As such, it is not clear how complex the generation task is, and we first try to use perplexity to describe this.

We train both \rnn models until \dev perplexity stops improving.
Our basic sequence-to-sequence model (\stos) reaches perplexity of 2.82 on \train and 2.92 on \dev after 15,000 batches of stochastic gradient descent.
The autoencoding sequence-to-sequence model  (\stosae) takes longer to fit, but reaches a lower minimum perplexity of 2.39 on \train and 2.51 on \dev after 25,000 batches.

To help ground perplexity numbers and understand the complexity of sentence biographies we train a benchmark language model and evaluate perplexity on \dev.
Following \newcite{LeBret16}, we build Kneser-Ney smoothed 5-gram language models using the KenLM toolkit \cite{Heafield:2011:kenlm}.

Table \ref{tab:templatedppls} lists perplexity numbers for the benchmark LM models with different templating schemes on \dev.
We observe decreasing perplexity for data with greater fact value templating.
\textsc{title} indicates templating of entity names only, while \textsc{full} indicates templating of all fact values by token index as described in \newcite{LeBret16}.
This shows that templating is an effective way to reduce the sparsity of a task, and that titles account for a large component of this.

Although \newcite{LeBret16} evaluate on a different dataset, we are able to draw some comparisons given the similarity of our task.
On their data, the benchmark LM baseline achieves a similar perplexity of 10.5 to ours when following their templating scheme on our dataset - suggesting both samples are of comparable complexity.

\begin{table}
    \centering
    \begin{tabular}{lcc}
        \hline
        Templates & \dev \\
        \hline \hline
        None & 29.8 \\ %79 \\
        Title & 14.5 \\ %53 \\
        Full & 10.1 \\ %09 \\
        \hline
    \end{tabular}
    \caption{Language model perplexity across templated datasets.}
    \label{tab:templatedppls}
\end{table}

\section{Model}
\label{sec-model}

We model the task as a sequence-to-sequence learning problem.
In this setting, a variable length input sequence of entity facts is encoded by a multi-layer \rnn into a fixed-length distributed representation.
This input representation is then fed into a separate decoder network which estimates a distribution over tokens as output.
During training, parameters for both the encoder and decoder networks are optimized to maximize the likelihood of a summary sequence given an observed fact sequence.

\begin{figure}
  \centering
  \includegraphics[width=\columnwidth]{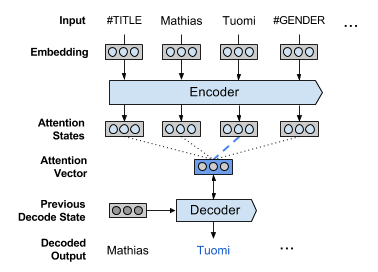}
  \caption{Sequence-to-sequence translation from linearized facts to text.}
  \label{fig-seq-model}
\end{figure}

Our setting differs from the translation task in that the input is a sequence representation of structured data rather than natural human language.
As described above in Section \ref{section:task}, we map Wikidata facts to a sequence of tokens that serves as input to the model as illustrated at the top of
Figure \ref{fig-seq-model}.
Experiments below demonstrate that this is sufficient for end-to-end learning in the generation task addressed here.
To generate summaries, our model must both select relevant content and transform it into a well formed sentence.
The decoder network includes an attention mechanism \cite{NIPS2015_5635} to help facilitate accurate content selection.
This allows the network to focus on different parts of the input sequence during inference.

\subsection{Sequence-to-sequence model (\stos)}

To generate language, we seed the decoder network with the output of the encoder and a designated \texttt{GO} token.
We then generate symbols greedily, taking the most likely output token from the decoder at each step given the preceding sequence until an \texttt{EOS} token is produced.
This approach follows \cite{DBLP:conf/nips/SutskeverVL14} who demonstrate a larger model with greedy sequence inference performs comparably to beam search.
In contrast to translation, we might expect good performance on the summarization task where output summary sequences tend to be well structured and often formulaic.
Additionally, we expect a partially-shared language across input and output.
To exploit this, we use a tied embedding space, which allows both the encoder and decoder networks to share information about word meaning between fact values and output tokens.

Our model uses a 3-layer stacked Gated Recurrent Unit \rnn for both encoding and decoding, implemented using TensorFlow.\footnote{\url{https://www.tensorflow.org}, v0.8.}
We limit the shared vocabulary to 100,000 tokens with 256 dimensions for each token embedding and hidden layer.
Less common tokens are marked as \texttt{UNK}, or unknown. %, making generated text less fluent.
To account for the long tail of entity names, we replace matches of title tokens with templated copy actions (e.g. \txt{TITLE0 TITLE1\ldots}).
These template are then filled after generation, as well as any initial unknown tokens in the output, which we fill with the first title token.
We learn using minibatch Stochastic Gradient Descent with a batch size of 64 and a fixed learning rate of 0.5.

\subsection{\stos with autoencoding (\stosae)}

One challenge for vanilla sequence-to-sequence models in this setting is the lack of a mechanism for constraining output sequences to only express those facts present in the data.
Given a fact extraction oracle, we might compare facts expressed in the output sequence with those of the input and appropriately adjust the loss for each instance.
While a forward-only model is only constrained to generate text sequences predicted by the facts, an autoencoding model is additionally constrained to generate text predictive of the input facts.
In place of this ideal setting, we introduce a second sequence-to-sequence model which runs in reverse - re-encoding the text output sequence of the forward model into facts.

\begin{figure}
  \centering
  \includegraphics[width=\columnwidth]{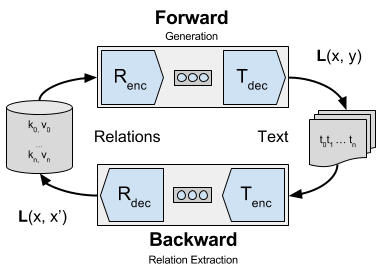}
  \caption{Sequence-to-sequence autoencoder.}
  \label{fig-model}
\end{figure}

This closed-loop model is detailed in Figure \ref{fig-model}.
The resulting network is trained end-to-end to minimize both the input-to-output sequence loss $L(x,y)$ and output-to-input reconstruction loss $L(x,x')$.
While gradients cannot propagate through the greedy forward decode step, shared parameters between the forward and backward network are fit to both tasks.
To generate language at test time, the backward network does not need to be evaluated.

\section{Experimental methodology}

The evaluation suite here includes standard baselines for comparison, automated metrics for learning, human judgement for evaluation and detailed analysis for diagnostics.
While each are individually useful, their combination gives a comprehensive analysis of a complex problem space.

\subsection{Benchmarks}

\paragraph{\wikitext}
We use the first sentence from Wikipedia both as a gold standard reference for evaluating generated sentences, and as an upper bound in human preference evaluation.

\paragraph{\base}
Template-based systems are strong baselines, especially in human evaluation.
While output may be stilted, the corresponding consistency can be an asset when consistency is important.
We induce common patterns from the \train set, replacing full matches of values with their slot and choosing randomly on ties.
Multiple non-fact tokens are collapsed to a single symbol.
A small sample of the most frequent patterns
were manually examined to produce templates, roughly expressed as: \txt{TITLE, known as GIVEN\_NAME, (born DATE\_OF\_BIRTH in PLACE\_OF\_BIRTH; died DATE\_OF\_DEATH in PLACE\_OF\_DEATH) is an POSITION\_HELD and OCCUPATION from CITIZENSHIP}, with some sensible back-offs where slots are not present, and rules for determiner agreement and \txt{is} versus \txt{was} where a death date is present.
For example, \txt{ollie freckingham (born 12 november 1988) is a cricketer from the united kingdom.}
In total, there are 48 possible template variations.

\subsection{Metrics}

\paragraph{\bleu}
We also report \bleu n-gram overlap with respect to the reference Wikipedia summary.
With a large dev/test sets (10,000 sentences here), \bleu is a reasonable evaluation of generated content.
However, it does not give an indication of well-formedness or readability.
Thus we complement \bleu with a human preference evaluation.

\paragraph{Human preference}
We use crowd-sourced judgements to evaluate the relative quality of generated sentences and the reference Wikipedia first sentence.
We obtain pairwise judgements, showing output from two different systems to crowd workers and asking each to give their binary preference.
The system name mappings are anonymized and ordered pseudo-randomly.
We request 3 judgements and dynamically increase this until we reach at least 70\% agreement or a maximum of 5 judgements.
We use CrowdFlower\footnote{\url{http://www.crowdflower.com}} to collect judgements at the cost of 31 USD for all 6 pairwise combinations over 82 randomly selected entities.
67 workers contributed judgements to the test data task, each providing no more than 50 responses.
We use the majority preference for each comparison.
The CrowdFlower agreement is 80.7\%, indicating that roughly 4 of 5 votes agree on average.

\subsection{Analysis of content selection}

Finally, no system is perfect, and it can be challenging to understand the inherent difficulty of the problem space and the limitations of a system.
Due to the limitations of the evaluation metrics mentioned above, we propose that manual annotation is important and still required for qualitative analysis to guide system improvement.
The structured data in knowledge-to-text tasks allows us, if we can identify expressions of facts in text, cases where facts have been omitted, incorrectly mentioned, or expressed differently.

\section{Results}
\label{sec-results}

\subsection{Comparison against Wikipedia reference}

Table \ref{tbl-bleu} shows \bleu scores calculated over 10,000 entities sampled from \dev and \test using the Wikipedia sentence as a single reference, using uniform weights for 1- to 4-grams, and padding sentences with fewer than 4 tokens.
Scores are similar across \dev and \test, indicating that the samples are of comparable difficulty.
We evaluate significance using bootstrapped resampling with 1,000 samples.
Each system result lies outside the 95\% confidence intervals of other systems.
\base has reasonable scores at 21, with \stos higher at around 32, indicating that the model is at least able to generate closer text than the baseline.
\stosae scores higher still at around 41, roughly double the baseline scores, indicating that the autoencoder is indeed able to constrain the model to generate better text.

\begin{table}
    \centering
    \begin{tabular}{lcc}
        \hline
         &  \dev & \test \\
        \hline \hline
        Base    & 21.3 & 21.1 \\
        \stos       & 32.5 & 33.1 \\
        \stosae     & 40.5 & 41.0 \\
        \hline
    \end{tabular}
    \caption{\bleu scores for each hypothesis against the Wikipedia reference}
    \label{tbl-bleu}
\end{table}

\subsection{Human preference evaluation}

Table \ref{fig:emnlp.test.agg.preferences} shows the results of our human evaluation over 82 entities sampled from \test.
For each pair of systems, we show the percentage of entities where the crowd preferred A over B.
Significant differences are annotated with $\ast$ and $\ast\ast$ for $p$ values $<$ 0.05 and 0.01 using a one-way $\chi^2$ test.
\wikitext is uniformly preferred to any system, as is appropriate for an upper bound.
The \stos model is the least-preferred with respect to \wikitext.
The \stosae model is more-preferred than the \base and \stos models, by a larger margin for the latter.
These results show that without autoencoding, the sequence-to-sequence model is less effective than a template-based system.
Finally, although \wikitext is more preferred than \stosae, the distributions are not significantly different, which we interpret as evidence that the model is able to generate good text from the human point-of-view, but autoencoding is required to do so.

\begin{table}
    \centering
    \begin{tabular}{p{1.2cm}p{1.2cm}p{1.2cm}|p{1cm}}
    \stosae & \base & \stos & \\
    \hline
    60\% & 61\%* & 87\%** & \wikitext \\
    & 62\%* & 77\%** & \stosae \\
    & & 65\%** & \base \\
    \end{tabular}
    \caption{Percentage of entities for which human judges preferred the row system to the column system. E.g., \stosae summaries are preferred to \base for 62\% of sample entities.}
    \label{fig:emnlp.test.agg.preferences}
\end{table}

\section{Analysis}
\label{sec-analysis}

While results presented above are encouraging and suggest that the model is performing well, they are not diagnostic in the sense that they can drive deeper insights into model strengths and weaknesses.
While inspection and manual analysis is still required, we also leverage the structured factual data inherent to our task to perform quantitative as well as qualitative analysis.

\subsection{Fact Count}

\begin{figure}
  \centering
  \includegraphics[width=.6\columnwidth]{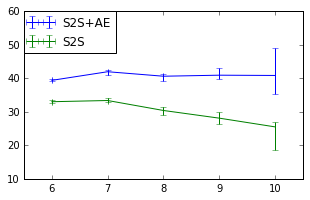}
  \caption{\bleu vs Fact Count on instances from DEV. Error bars indicate the 95\% confidence interval for \bleu.}
  \label{fig-bleu-vs-num-rels}
\end{figure}

Figure \ref{fig-bleu-vs-num-rels} shows the effects of input fact count on generation performance.
While more input facts give more information for the model to work with, longer inputs are also both rarer and more complex to encode.
Interestingly, we observe the \stosae model maintains performance for more complex inputs while \stos performance declines.

\begin{table*}[t!]
    \centering
  \begin{tabular}{p{1.2cm}p{1cm}p{12.5cm}}
  \hline \hline
  Data & & COUNTRY\_OF\_CITIZENSHIP united states of america DATE\_OF\_BIRTH 16/04/1927 DATE\_OF\_DEATH 19/05/1959 OCCUPATION formula one driver PLACE\_OF\_BIRTH redlands PLACE\_OF\_DEATH indianapolis SEX\_OR\_GENDER male TITLE bob cortner \\
\hline
\wikitext & n/a & \txt{\textbf{robert} \textit{charles} \textbf{cortner} ( \textbf{april 16 , 1927} – \textbf{may 19 , 1959} ) was an \textbf{american} \textbf{automobile racing driver} from \textit{redlands , california} .} \\
\base & 47.7 & \txt{\textbf{bob cortner} ( born \textbf{16 april 1927} in \textbf{redlands} ; died \textbf{19 may 1959} in \textbf{indianapolis} ) was a \textbf{formula one driver} from the \textbf{united states of america}} \\
\stos & 45.7 & \txt{\textbf{bob cortner} ( \textbf{april 16 , 1927} – \sout{may 19 , 2005} ) was an \textbf{american} \sout{professional boxer} .} \\
\stosae & 58.8 & \txt{\textbf{robert cortner} ( \textbf{april 16 , 1927} – \textbf{may 19 , 1959} ) was an \textbf{american} \textbf{racecar driver} .} \\

\hline \hline
Data & & COUNTRY\_OF\_CITIZENSHIP united kingdom DATE\_OF\_BIRTH 08/01/1906 DATE\_OF\_DEATH 12/12/1985 OCCUPATION actor PLACE\_OF\_BIRTH london PLACE\_OF\_DEATH chelsea SEX\_OR\_GENDER male TITLE barry mackay (actor) \\
\hline
\wikitext & n/a & \txt{\textbf{barry mackay} ( \textbf{8 january 1906} – \textbf{12 december 1985} ) was a \textbf{british} \textbf{actor}.} \\
\base & 34.3 & \txt{\textbf{barry mackay ( actor )} ( born \textbf{8 january 1906} in \textbf{london} ; died \textbf{12 december 1985 in chelsea} ) was an \textbf{actor} from the \textbf{united kingdom} .} \\
\stos & 84.8 & \txt{\textbf{barry mackay} ( \textbf{8 january 1906} – \textbf{12 december 1985} ) was a \textbf{british} \sout{film actor} .} \\
\stosae & 76.7 & \txt{\textbf{barry mackay} ( \textbf{8 january 1906} – \textbf{12 december 1985} ) was an \textbf{english} \textbf{actor} .} \\

\hline \hline
Data & & COUNTRY\_OF\_CITIZENSHIP united states of america DATE\_OF\_BIRTH 27/08/1931 DATE\_OF\_DEATH 03/11/1995 OCCUPATION jazz musician SEX\_OR\_GENDER male TITLE joseph "flip" nu{\~n}ez \\
\hline
\wikitext & n/a & \txt{\textbf{joseph} \textit{`` flip '} \textbf{nu{\~n}ez} was an \textbf{american} \sout{jazz pianist} , \sout{composer} , and \sout{vocalist} of \textit{filipino} descent .} \\
\base & 15.0 & \txt{\textbf{joseph} \textit{`` flip '} \textbf{nu{\~n}ez} ( born \textbf{27 august 1931} ; died \textbf{3 november 1995} ) was a \textbf{jazz musician} from the \textbf{united states of america} .} \\
\stos & 29.1 & \txt{\textbf{joseph} \textit{`` flip '} \textbf{nu{\~n}ez} ( \textbf{august 27 , 1931} – \textbf{november 3 , 1995} ) was an \textbf{american} \sout{jazz trumpeter} .} \\
\stosae & 29.1 & \txt{\textbf{joseph} \textit{`` flip '} \textbf{nu{\~n}ez} ( \textbf{august 27 , 1931} – \textbf{november 3 , 1995} ) was an \textbf{american} \sout{jazz drummer} .} \\
  \hline \hline
  \end{tabular}
  \caption{Examples of entities from \dev, showing facts, \wikitext, \base, \stos and \stosae. %We show the \bleu score against the \wiki reference for all system output. 
  We mark \textbf{correct}, \sout{incorrect} and \textit{extra} fact values in the text with respect to the Wikidata input.}
  \label{example-output}
\end{table*}

\subsection{Example generated text}

Table \ref{example-output} shows some \dev entities and their summaries.
The model learns interesting mappings: between numeric and string dates, and country demonyms.
The model also demonstrates the ability to work around edge cases where templates fail, i.e. stripping parenthetical disambiguations (e.g. \txt{(actor)}) and emitting the name \txt{Robert}
when the input is \txt{Bob}.
Output also suggests the model may perform inference across multiple facts to improve generation precision, e.g. describing an entity as \txt{english} rather than \txt{british} given information about both citizenship and place of birth.
Unfortunately, the model can also infer unsubstantiated facts into the text (i.e. \txt{jazz drummer}).

\subsection{Content selection and hallucination}
\label{sec:analysis}

We randomly sample 50 entities from \dev and manually annotate the Wikipedia and system text. We note which fact slots are expressed as well as whether the expressed values are correct with respect to Wikidata.
Given two sets of correctly extracted facts, we can consider one \emph{gold}, one \emph{system} and calculate set-based precision, recall and F1.

\paragraph{What percentage of facts are used in the reference summaries?}
Firstly, to understand how Wikipedia editors select content for the first sentence of articles, we measure recall with the real facts as gold, and Wikipedia as system.
Overall, the recall is 0.61 indicating that 61\% of input facts are expressed in the reference summary from Wikipedia.
The entity name (TITLE) is always expressed.
Four slots are nearly always expressed when available: OCCUPATION (90\%), DATE\_OF\_BIRTH (84\%), CITIZENSHIP (81\%), DATE\_OF\_DEATH (80\%).
Six slots are infrequently expressed in the analysis sample: PLACE\_OF\_BIRTH (33\%), POSITION\_HELD (25\%), PARTICIPANT\_OF (20\%), POLITICAL\_PARTY (20\%), EDUCATED\_AT (14\%), SPORTS\_TEAM (9\%).
Two are never expressed explicitly: PLACE\_OF\_DEATH (0\%), SEX\_OR\_GENDER (0\%).
AWARD\_RECEIVED and SPORT are not in the analysis sample.

\paragraph{Do systems select the same facts found in the reference summaries?}
Table \ref{tab:factprf} shows content selection scores for systems with respect to the Wikipedia text as reference.
This suggests that the autoencoding in \stosae helps increase fact recall without sacrificing precision.
The template baseline also attains this higher recall, but at the cost of precision.
For commonly expressed facts found in most person biographies, recall is over 0.95 (e.g., CITIZENSHIP, BIRTH\_DATE, DEATH\_DATE and OCCUPATION).
Facts that are infrequently expressed are more difficult to select, with system F1 ranging from 0.00 to 0.50.
Interestingly, macro-averaged F1 across infrequently expressed facts mirror human preference rather than \bleu results, with \stosae (0.26) $>$ \base (0.17) $>$ \stos (0.07).
However, all systems perform poorly on these facts and no reliable differences are observed.

 \begin{table}
     \centering
     \begin{tabular}{lrrr}
         \hline
          & P & R & F \\
         \hline \hline
         \base & 0.80 & 0.79 & 0.79 \\
         \stos & 0.89 & 0.67 & 0.77 \\
         \stosae & 0.89 & 0.78 & 0.83 \\
         \hline
     \end{tabular}
     \caption{Fact-set content selection results phrased as precision, recall and F1 of systems with respect to the Wikipedia reference on \dev.}
     \label{tab:factprf}
 \end{table}

\paragraph{How does autoencoding effect fact density?}
Interestingly, we observe that the autoencoding objective encourages the model to select more facts (5.2 for \stosae vs.\ 4.5 for \stos), without increasing sentence length (19.1 vs.\ 19.7 tokens).
\base is similarly productive (5.1 facts) but wordier (21.2 tokens), while the \wikitext reference produces both more facts (6.1) and longer sentences (23.7).

\paragraph{Do systems hallucinate facts?}
To quantify the effect of hallucinated facts, we asses content selection scores of systems with respect to the input Wikidata relations (Table \ref{tab:hallucprf}).
Our best model achieves a precision of 0.93 with respect to Wikidata input.
Notably, the template-driven baseline maintains a precision of 1.0 as it is constrained to emit Wikidata facts verbatim.

 \begin{table}
     \centering
     \begin{tabular}{lrrr}
         \hline
          & P & R & F \\
         \hline \hline
         \base & 1.00 & 0.74 & 0.85 \\
         \stos & 0.96 & 0.55 & 0.70 \\
         \stosae & 0.93 & 0.62 & 0.74 \\
         \wikitext & 0.81 & 0.61 & 0.69 \\
         \hline
     \end{tabular}
     \caption{Hallucination results phrased as precision, recall and F1 of systems with respect to the Wikidata input on \dev.}
     \label{tab:hallucprf}
 \end{table}

\section{Discussion and future work}

Our experiments show that \rnns can generate biographic summaries from structured data, and that a secondary autoencoding objective is able to account for some of the information mismatch between input facts and target output sentences.
In the future, we will explore whether results improve with explicit modelling of facts and conditioning of generation and autoencoding losses on slots.
We expect this could benefit generation for diverse and noisy slot schemas like Wikipedia Infoboxes.

Another natural extension is to investigate the performance of the network running in reverse, from summary text back to facts.
We plan to isolate the performance of the \stosae backward model when inferring facts and compare it to standard relation extraction systems.
Finally, similar \rnn models have been applied extensively to language translation tasks.
We plan to explore whether a joint model of machine translation and fact-driven generation can help populate \kb entries for low-coverage languages by leveraging a shared set of facts.

\section{Conclusion}

We present a neural model for mapping between structured and unstructured data, focusing on creating Wikipedia biographic summary sentences from Wikidata slot-value pairs.
We introduce a sequence-to-sequence autoencoding \rnn which improves upon base models by jointly learning to generate text and reconstruct facts.
Our analysis of the task suggests evaluation in this domain is challenging.
In place of a single score, we analyse statistical measures, human preference judgements and manual annotation to help characterise the task and understand system performance.
In the human preference evaluation, our best model outperforms template baselines and is preferred 40\% of the time to the gold standard Wikipedia reference.

Code and data is available at
\url{https://github.com/andychisholm/mimo}.

\section*{Acknowledgments}
This work was supported by a Google Faculty Research Award (Chisholm) and an Australian Research Council Discovery Early Career Researcher Award (DE120102900, Hachey).
Many thanks to reviewers for insightful comments and suggestions, and to Glen Pink, Kellie Webster, Art Harol and Bo Han for feedback at various stages.

%\clearpage
\bibliography{wiki-bio-summ}

\begin{thebibliography}{}

\bibitem[\protect\citename{Angeli \bgroup et al.\egroup
  }2010]{angeli-liang-klein:2010:EMNLP}
Gabor Angeli, Percy Liang, and Dan Klein.
\newblock 2010.
\newblock A simple domain-independent probabilistic approach to generation.
\newblock In {\em Conference on Empirical Methods in Natural Language
  Processing}, pages 502--512.

\bibitem[\protect\citename{Bamman and Smith}2014]{DBLP:journals/tacl/BammanS14}
David Bamman and Noah~A. Smith.
\newblock 2014.
\newblock Unsupervised discovery of biographical structure from text.
\newblock {\em Transactions of the Association for Computational Linguistics},
  2:363--376.

\bibitem[\protect\citename{Biadsy \bgroup et al.\egroup
  }2008]{biadsy-hirschberg-filatova:2008:ACLMain}
Fadi Biadsy, Julia Hirschberg, and Elena Filatova.
\newblock 2008.
\newblock An unsupervised approach to biography production using {Wikipedia}.
\newblock In {\em Annual Meeting of the Association for Computational
  Linguistics}, pages 807--815.

\bibitem[\protect\citename{Bird \bgroup et al.\egroup }2009]{BirdKleinLoper09}
Steven Bird, Ewan Klein, and Edward Loper.
\newblock 2009.
\newblock {\em Natural Language Processing with Python: Analyzing Text with the
  Natural Language Toolkit}.
\newblock O'Reilly Media.

\bibitem[\protect\citename{{Blair-Goldensohn} \bgroup et al.\egroup
  }2004]{Blair:2004}
Sasha {Blair-Goldensohn}, David Evans, Vasileios Hatzivassiloglou, Kathleen
  McKeown, Ani Nenkova, Rebecca Passonneau, Barry Schiffman, Andrew Schlaikjer,
  Advaith Siddharthan, and Sergey Siegelman.
\newblock 2004.
\newblock {Columbia University} at {DUC} 2004.
\newblock In {\em Proceedings of the Document Understanding Workshop}, pages
  23--30.

\bibitem[\protect\citename{Cheng \bgroup et al.\egroup
  }2015]{Cheng:2015:SED:2736277.2741094}
Gong Cheng, Danyun Xu, and Yuzhong Qu.
\newblock 2015.
\newblock Summarizing entity descriptions for effective and efficient
  human-centered entity linking.
\newblock In {\em International Conference on World Wide Web}, pages 184--194.

\bibitem[\protect\citename{Dai and Le}2015]{DBLP:conf/nips/DaiL15}
Andrew~M. Dai and Quoc~V. Le.
\newblock 2015.
\newblock Semi-supervised sequence learning.
\newblock In {\em Annual Conference on Neural Information Processing Systems},
  pages 3079--3087.

\bibitem[\protect\citename{Duboue and McKeown}2003]{Duboue-McKeown:2003:EMNLP}
Pablo~Ariel Duboue and Kathleen~R McKeown.
\newblock 2003.
\newblock Statistical acquisition of content selection rules for natural
  language generation.
\newblock In {\em Conference on Empirical Methods in Natural Language
  Processing}, pages 121--128.

\bibitem[\protect\citename{Duma and Klein}2013]{duma-klein:2013:IWCS2013}
Daniel Duma and Ewan Klein.
\newblock 2013.
\newblock Generating natural language from linked data: Unsupervised template
  extraction.
\newblock In {\em International Conference on Computational Semantics}, pages
  83--94.

\bibitem[\protect\citename{Garera and Yarowsky}2009]{garera-yarowsky:2009:EACL}
Nikesh Garera and David Yarowsky.
\newblock 2009.
\newblock Structural, transitive and latent models for biographic fact
  extraction.
\newblock In {\em Conference of the European Chapter of the Association for
  Computational Linguistics}, pages 300--308.

\bibitem[\protect\citename{Gillick \bgroup et al.\egroup
  }2016]{gillick-EtAl:2016:N16-1}
Dan Gillick, Cliff Brunk, Oriol Vinyals, and Amarnag Subramanya.
\newblock 2016.
\newblock Multilingual language processing from bytes.
\newblock In {\em Conference of the North American Chapter of the Association
  for Computational Linguistics}, pages 1296--1306.

\bibitem[\protect\citename{Gyawali and
  Gardent}2014]{gyawali-gardent:2014:P14-1}
Bikash Gyawali and Claire Gardent.
\newblock 2014.
\newblock Surface realisation from knowledge-bases.
\newblock In {\em Annual Meeting of the Association for Computational
  Linguistics}, pages 424--434.

\bibitem[\protect\citename{Han \bgroup et al.\egroup
  }2015]{han-EtAl:2015:W15-46}
Sangdo Han, Jeesoo Bang, Seonghan Ryu, and Gary~Geunbae Lee.
\newblock 2015.
\newblock Exploiting knowledge base to generate responses for natural language
  dialog listening agents.
\newblock In {\em Annual Meeting of the Special Interest Group on Discourse and
  Dialogue}, pages 129--133.

\bibitem[\protect\citename{Heafield}2011]{Heafield:2011:kenlm}
Kenneth Heafield.
\newblock 2011.
\newblock {KenLM}: Faster and smaller language model queries.
\newblock In {\em Workshop on Statistical Machine Translation}, pages 187--197.

\bibitem[\protect\citename{Kondadadi \bgroup et al.\egroup
  }2013]{kondadadi-howald-schilder:2013:ACL2013}
Ravi Kondadadi, Blake Howald, and Frank Schilder.
\newblock 2013.
\newblock A statistical {NLG} framework for aggregated planning and
  realization.
\newblock In {\em Annual Meeting of the Association for Computational
  Linguistics}, pages 1406--1415.

\bibitem[\protect\citename{Konstas and
  Lapata}2012]{konstas-lapata:2012:ACL2012}
Ioannis Konstas and Mirella Lapata.
\newblock 2012.
\newblock Concept-to-text generation via discriminative reranking.
\newblock In {\em Annual Meeting of the Association for Computational
  Linguistics}, pages 369--378.

\bibitem[\protect\citename{Lebret \bgroup et al.\egroup }2016]{LeBret16}
R{\'{e}}mi Lebret, David Grangier, and Michael Auli.
\newblock 2016.
\newblock Neural text generation from structured data with application to the
  biography domain.
\newblock In {\em Conference on Empirical Methods in Natural Language
  Processing}, pages 1203--1213.

\bibitem[\protect\citename{Lin and Hovy}2003]{lin-naacl03-automatic}
Chin-Yew Lin and Eduard Hovy.
\newblock 2003.
\newblock Automatic evaluation of summaries using n-gram co-occurrence
  statistics.
\newblock In {\em Conference of the North American Chapter of the Association
  for Computational Linguistics}, pages 71--78.

\bibitem[\protect\citename{Luong \bgroup et al.\egroup
  }2016]{DBLP:journals/corr/LuongLSVK15}
Minh{-}Thang Luong, Quoc~V. Le, Ilya Sutskever, Oriol Vinyals, and Lukasz
  Kaiser.
\newblock 2016.
\newblock Multi-task sequence to sequence learning.
\newblock In {\em International Conference on Learning Representations}.

\bibitem[\protect\citename{Mei \bgroup et al.\egroup }2015]{mei2015selective}
Hongyuan Mei, Mohit Bansal, and Matthew~R. Walter.
\newblock 2015.
\newblock What to talk about and how? {Selective} generation using {LSTMs} with
  coarse-to-fine alignment.
\newblock In {\em Conference of the North American Chapter of the Association
  for Computational Linguistics}, pages 720--730.

\bibitem[\protect\citename{Nenkova and
  Passonneau}2004]{nenkova-passonneau:2004:HLTNAACL}
Ani Nenkova and Rebecca Passonneau.
\newblock 2004.
\newblock Evaluating content selection in summarization: The pyramid method.
\newblock In {\em Conference of the North American Chapter of the Association
  for Computational Linguistics}, pages 145--152.

\bibitem[\protect\citename{Papineni \bgroup et al.\egroup
  }2002]{Papineni:2002:BMA:1073083.1073135}
Kishore Papineni, Salim Roukos, Todd Ward, and Wei-Jing Zhu.
\newblock 2002.
\newblock {BLEU}: A method for automatic evaluation of machine translation.
\newblock In {\em Annual Meeting on Association for Computational Linguistics},
  pages 311--318.

\bibitem[\protect\citename{Power and Third}2010]{power-third:2010:POSTERS}
Richard Power and Allan Third.
\newblock 2010.
\newblock Expressing {OWL} axioms by english sentences: Dubious in theory,
  feasible in practice.
\newblock In {\em International Conference on Computational Linguistics}, pages
  1006--1013.

\bibitem[\protect\citename{Rush \bgroup et al.\egroup
  }2015]{rush-chopra-weston:2015:EMNLP}
Alexander~M. Rush, Sumit Chopra, and Jason Weston.
\newblock 2015.
\newblock A neural attention model for abstractive sentence summarization.
\newblock In {\em Conference on Empirical Methods in Natural Language
  Processing}, pages 379--389.

\bibitem[\protect\citename{Schiffman \bgroup et al.\egroup
  }2001]{schiffman-mani-concepcion:2001:ACL}
Barry Schiffman, Inderjeet Mani, and Kristian Concepcion.
\newblock 2001.
\newblock Producing biographical summaries: Combining linguistic knowledge with
  corpus statistics.
\newblock In {\em Annual Meeting of the Association for Computational
  Linguistics}, pages 458--465.

\bibitem[\protect\citename{Sutskever \bgroup et al.\egroup
  }2014]{DBLP:conf/nips/SutskeverVL14}
Ilya Sutskever, Oriol Vinyals, and Quoc~V. Le.
\newblock 2014.
\newblock Sequence to sequence learning with neural networks.
\newblock In {\em Annual Conference on Neural Information Processing Systems},
  pages 3104--3112.

\bibitem[\protect\citename{Vinyals and Le}2015]{DBLP:journals/corr/VinyalsL15}
Oriol Vinyals and Quoc~V. Le.
\newblock 2015.
\newblock A neural conversational model.
\newblock In {\em ICML Deep Learning Workshop}.

\bibitem[\protect\citename{Vinyals \bgroup et al.\egroup }2015]{NIPS2015_5635}
Oriol Vinyals, {\L}ukasz Kaiser, Terry Koo, Slav Petrov, Ilya Sutskever, and
  Geoffrey Hinton.
\newblock 2015.
\newblock Grammar as a foreign language.
\newblock In {\em Annual Conference on Neural Information Processing Systems},
  pages 2755--2763.

\bibitem[\protect\citename{Wen \bgroup et al.\egroup
  }2015]{wen-EtAl:2015:EMNLP}
Tsung-Hsien Wen, Milica Gasic, Nikola Mrk\v{s}i\'{c}, Pei-Hao Su, David
  Vandyke, and Steve Young.
\newblock 2015.
\newblock Semantically conditioned {LSTM}-based natural language generation for
  spoken dialogue systems.
\newblock In {\em Conference on Empirical Methods in Natural Language
  Processing}, pages 1711--1721.

\bibitem[\protect\citename{Xiao \bgroup et al.\egroup
  }2016]{xiao-dymetman-gardent:2016:P16-1}
Chunyang Xiao, Marc Dymetman, and Claire Gardent.
\newblock 2016.
\newblock Sequence-based structured prediction for semantic parsing.
\newblock In {\em Annual Meeting of the Association for Computational
  Linguistics}, pages 1341--1350.

\bibitem[\protect\citename{Zhang \bgroup et al.\egroup
  }2012]{Zhang:2012:SHS:2348283.2348306}
Lanbo Zhang, Yi~Zhang, and Yunfei Chen.
\newblock 2012.
\newblock Summarizing highly structured documents for effective search
  interaction.
\newblock In {\em International Conference on Research and Development in
  Information Retrieval}, pages 145--154.

\end{thebibliography}
\bibliographystyle{eacl2017}

\end{document}